\documentclass[conference]{IEEEtran}
\IEEEoverridecommandlockouts
\usepackage{cite}
\usepackage{amsmath,amssymb,amsfonts}

\usepackage{algorithmic}
\usepackage{graphicx}
\usepackage{textcomp}
\usepackage{xcolor}
\usepackage{enumerate}
\usepackage{enumitem}
\usepackage{amsmath}
\usepackage{mathabx}
\usepackage{float}
\usepackage{algorithm}
\usepackage{algorithmic}
\usepackage{multirow}
\usepackage{dsfont}
\usepackage{float}
\usepackage{multirow}
\usepackage[normalem]{ulem}
\usepackage{url}

\usepackage{bm}
\usepackage{subfigure}
\usepackage{caption}
\usepackage{nicefrac}

\usepackage{array}
\newcolumntype{L}[1]{>{\raggedright\let\newline\\\arraybackslash\hspace{0pt}}m{#1}}
\newcolumntype{C}[1]{>{\centering\let\newline  \\\arraybackslash\hspace{0pt}}m{#1}}
\newcolumntype{R}[1]{>{\raggedleft\let\newline \\\arraybackslash\hspace{0pt}}m{#1}}

\newtheorem{Definition}{Definition}

\def\BibTeX{{\rm B\kern-.05em{\sc i\kern-.025em b}\kern-.08em
	T\kern-.1667em\lower.7ex\hbox{E}\kern-.125emX}}
\begin{document}

\title{Efficient Relation-aware Scoring Function Search
	for Knowledge Graph Embedding}

\author{\IEEEauthorblockN{Shimin DI$^1$, Quanming YAO$^{2,3}$, Yongqi ZHANG$^{2}$, Lei CHEN$^1$}
	$^1$The Hong Kong University of Science and Technology, Hong Kong SAR, China\\
	$^2$4Paradigm Inc. Hong Kong SAR, China\\
	$^3$Department of Electronic Engineering, Tsinghua University, Beijing, China\\
	\{sdiaa,leichen\}@cse.ust.hk, \{yaoquanming,zhangyongqi\}@4paradigm.com}

\maketitle

\begin{abstract}
The scoring function, which measures the plausibility of triplets in knowledge graphs (KGs), 
is the key to ensure the excellent performance of KG embedding, and its design is also an important problem in the literature.
Automated machine learning (AutoML) techniques have recently been introduced into KG to design task-aware scoring functions, which achieve the state-of-the-art performance in KG embedding.
However, the effectiveness of searched scoring functions is still not as good as desired. 
In this paper, observing that existing scoring functions can exhibit distinct performance on different semantic patterns, we are motivated to explore such semantics by searching relation-aware scoring functions.
But the relation-aware search requires a much larger search space than the previous one.
Hence, we propose to encode the space as a supernet and propose an efficient alternative minimization algorithm to search through the supernet in a one-shot manner. 
Finally, experimental results on benchmark datasets demonstrate that the proposed method can efficiently search relation-aware scoring functions, and achieve better embedding performance than state-of-the-art methods.
\footnote{The work was done when S. Di was an intern in 4Paradigm Inc. mentored by Q. Yao;
	and Correspondence is to Q.Yao.}
\end{abstract}

\begin{IEEEkeywords}
	Knowledge Graph,
	Knowledge Graph Embedding,
	Neural Architecture Search,
	Automated Machine Learning
\end{IEEEkeywords}

\setcounter{section}{0}
\section{Introduction}
\label{sec:intro}
Knowledge Graph (KG) \cite{nickel2015review,wang2017knowledge}, 
as one of the most effective ways to explore and organize knowledge base, 
applies to various problems, such as question answering \cite{lukovnikov2017neural}, 
recommendation \cite{zhang2016collaborative}, and few-shot learning \cite{wang2018zero}.
In KGs, every edge represents a knowledge triplet in the form of $\textit{(head entity, relation, tail entity)}$, 
or $(h, r, t)$ for simplicity. 
Given a triplet, several crucial tasks in KGs, 
such as link prediction and triplet classification~\cite{nickel2015review,wang2017knowledge}, 
can be used to verify whether such a fact exists to form this triplet. 
KG embedding has been proposed to address this issue. 
Basically, 
KG embedding targets to embed entities $h,t$ and relations $r$ into low-dimensional vector space such as $\bm{h},\bm{r},\bm{t}\in \mathbb{R}^d$. 
Then based on the embeddings, 
a scoring function $f$ is employed to compute a score $f(\bm{h},\bm{r},\bm{t})$ 
to verify whether a triplet $(h, r, t)$ is a fact. 
Triplets with higher scores are more likely to be facts. 

In the last decade, various scoring functions have been proposed to significantly improve the quality of embeddings \cite{lin2018knowledge,nickel2015review,wang2017knowledge}.	
TransE~\cite{bordes2013translating}, as a representative scoring function, interprets the relation $r$ as a translation from head entity $h$ to tail entity $t$, and optimizes the embeddings to satisfy $\bm{h} + \bm{r} = \bm{t}$.
However, TransE~\cite{bordes2013translating} and its variants \cite{wang2014knowledge,lin2015learning} are not fully expressive and their empirical performance is inferior to the others as mentioned in~\cite{wang2018multi}.
Recently, some works~\cite{socher2013reasoning,dettmers2018convolutional,dong2014knowledge} employ neural networks to design universal scoring functions.
But these scoring functions are not well-regularized for the KG properties and are expensive to general predictions~\cite{zhang2019autosf}.
Furthermore, bilinear models (BLMs)~\cite{yang2014embedding,trouillon2017knowledge,kazemi2018simple,liu2017analogical,lacroix2018canonical} are proposed to compute the score by the weighted sum of pairwise interactions of embeddings.
Currently, BLMs are the most powerful in terms of both empirical results and theoretical guarantees~\cite{lacroix2018canonical} on expressiveness ~\cite{wang2018multi,kazemi2018simple}.

Generally,
the models in BLMs share the form as 
$f(\bm{h}, \bm{r}, \bm{t})=\bm{h}^\top \! g(\bm{r})\bm{t}$,
where $g(\bm{r})$ returns a square matrix referring to the relation $\bm{r}$.
DistMult~\cite{yang2014embedding} regularizes 
$g(\bm{r})$
to be diagonal, such as $g(\bm{r})=\text{diag}(\bm{r})$, in order to solve the overfitting problem.
ComplEx~\cite{trouillon2017knowledge} extends the embeddings to be complex values.
SimplE~\cite{kazemi2018simple} is another variant that regularizes the matrix $g(\bm{r})$
with a simpler but valid constraint.
More recently, 
TuckER~\cite{balazevic2019tucker} extends BLMs to
tensor models for KG embedding.
However, the designing of scoring functions is still challenging because of the diversity of KGs \cite{wang2017knowledge}.
A scoring function performs well on one task may not adapt well to the other tasks 
since different KGs usually own distinct patterns~\cite{zhang2019autosf}, especially the relation patterns~\cite{wang2017knowledge}.

Recently, automated machine learning (AutoML)~\cite{automl_book,yao2018taking},
as demonstrated via automated hyperparameter tuning~\cite{feurer2015efficient} and neural architecture search (NAS)~\cite{zoph2016neural}, has shown to be very useful in the design of machine learning models. 
Inspired by such a success, a pioneered work, AutoSF~\cite{zhang2019autosf}, has proposed to search an appropriate scoring function for any given KG data using AutoML techniques.
AutoSF first defines the manual scoring functions design problem as a scoring function search problem and then proposes a search algorithm.
It empirically shows that the searched scoring functions are KG dependent and outperform the state-of-the-art ones designed by human experts. 
In short, AutoSF can search for proper scoring functions, which depend on the given KG data and evaluation metric.
In other words, AutoSF is task-aware, while traditional scoring functions are not.

\setcounter{table}{0}
\renewcommand{\tablename}{TABLE}
\begin{table*}[ht]
	\caption{The summary of existing scoring functions.
		The expressiveness measures whether a scoring function can handle
		common patterns in KGs: symmetry, anti-symmetry, general asymmetry, inversion.
		We compare the inference cost on single triplet of scoring functions w.r.t. the embedding dimension $d$.
		$N_e$ and $N_r$ denote the number of entities and relations, respectively.}
	\centering
	\vspace{-5px}
	\label{tab:overview}
	\begin{tabular}{ c | c | c | c | c | c| c}
		\hline
		& \multirow{2}{*}{\bf Scoring functions} &    \multicolumn{3}{c|}{\bf Effectiveness}     & \multicolumn{2}{c}{\bf Efficiency} \\ \cline{3-7}
		&        & expressive & task-aware & relation-aware & inference time & model complexity \\ \hline
		& TransE \cite{bordes2013translating}    & $\times$        & $\times$   & $\times$       & $O(d)$       & $O(N_ed+N_rd)$ \\ \cline{2-7}
		Hand-designed & DistMult \cite{yang2014embedding}      & $\times$        & $\times$   & $\times$       & $O(d)$       &  $O(N_ed+N_rd)$ \\ \cline{2-7}
		& NTN~\cite{socher2013reasoning}         & $\surd$         & $\times$   & $\times$       & $O(d^2)$    & $O(N_ed+N_rd^2)$  \\ \cline{2-7}
		& TuckER  \cite{balazevic2019tucker}     & $\surd$         & $\times$   & $\times$       & $O(d^3)$   &   $O(d^3+N_ed+N_rd)$  \\ \cline{2-7}
		& ComplEx  \cite{trouillon2017knowledge} & $\surd$         & $\times$   & $\times$       & $O(d)$     &  $O(N_ed+N_rd)$  \\ \cline{2-7}
		& HypER~\cite{balavzevic2019hypernetwork}          & $\surd$         & $\times$   & $\times$       & $O(d^2)$    &   $O(N_ed+N_rd)$ \\ \hline
		Searched      & AutoSF \cite{zhang2019autosf}          & $\surd$         & $\surd$    & $\times$       & $O(d)$     &  $O(N_ed+N_rd)$   \\ \cline{2-7}
		(by AutoML)   & ERAS                                   & $\surd$         & $\surd$    & $\surd$        & $O(d)$     &  $O(N_ed+N_rd)$   \\ \hline
	\end{tabular}
\end{table*}

However,
the task-aware method AutoSF still follows the classic way that forcing all relations to share one scoring function.
This is not relation-aware and may cause the effectiveness issue.
Generally, common KG relations can be roughly categorized into different patterns based on their semantic properties, such as  symmetry~\cite{yang2014embedding}, anti-symmetry~\cite{trouillon2017knowledge,nickel2016holographic}, general asymmetry~\cite{liu2017analogical}, and inversion~\cite{kazemi2018simple}. 
Traditional models design universal scoring functions to cover more and more relation patterns. 
For example, DistMult~\cite{yang2014embedding} only handles symmetric relations, while TransE~\cite{bordes2013translating} covers another three kinds of relations except for symmetric relations. Furthermore, ComplEx~\cite{trouillon2017knowledge} and SimplE~\cite{kazemi2018simple} are able to cover all these four common relation patterns. 
Intuitively, the more patterns covered by the scoring function, the stronger ability it has to learn KGs.
However, there are some potential risks in the pursuit of universal scoring functions. 
A universal scoring function may not perform well on certain relation patterns even though it can handle all kinds of relations~\cite{rossi2020knowledge}. 
For instance, it has been reported in~\cite{meilicke2018fine}
that HolEX~\cite{xue2018expanding} achieves unsatisfactory performance on symmetric relations in the FB15K data set~\cite{bordes2013translating}, despite HolEX being a universal scoring function.
This indicates that forcing all relations to share one scoring function may not be able to fully express the interactions between entities with different relations. 
As compared in Table~\ref{tab:overview},
none of the existing methods cover all the aspects
in terms of the effectiveness.

Unfortunately, it is hard to extend the task-aware method (i.e., AutoSF~\cite{zhang2019autosf}) to be relation-aware due to the efficiency issue.
AutoSF adopts a progressive greedy search approach to find a universal scoring function.
It requires separately training hundreds of scoring functions to convergence, which suffers from a lot of computational overhead.
In general, AutoSF takes more than one GPU day to search on the smallest benchmark data set WN18RR, 
while it requires more than 9 GPU days on the larger data set YAGO.
But the relation-aware search problem requires a much larger search space than the space of AutoSF.
Therefore, a much more efficient search algorithm is needed for the relation-aware search.

In this paper, to address the issues mentioned above, we propose the \textit{E}fficient search method for \textit{R}elation-\textit{A}ware \textit{S}coring functions (ERAS) in KG embedding.
We propose to search multiple scoring functions, which are expressive, task-aware, and relation-aware, for common relation patterns in any given KG data. 
More concretely, we propose a supernet to model the relation-aware search space and introduce an efficient search approach to the supernet. 
We suggest sharing KG embeddings on the supernet, so as to avoid training hundreds of candidate scoring functions from scratch as AutoSF does.
In summary, we list the contribution we have made in this work as follows:

\begin{itemize}[leftmargin=*]
	\item 
	Previous works mainly emphasize the expressiveness of scoring functions, which also motivates AutoSF to design task-aware scoring functions.
	However, they ignore that scoring functions should also be relation-aware as they model the semantics of relations.
	In this paper, to address such a problem, we propose an AutoML-based method to design relation-aware scoring functions.
	
	\item 
	We define a novel supernet representation to model the relation-aware search space, where the relations are assigned into different groups and each group has a unique scoring function.
	The simple but effective supernet not only enables us to share KG embeddings to significantly accelerate the search but also protects our search from negative effects.
	
	\item 
	Inspired by the one-shot architecture search (OAS) algorithms,
	we propose a stochastic algorithm, i.e., ERAS, that is efficient and suitable for the automated scoring function search task.
	It optimizes the search problem through alternative minimization, where embeddings are stochastically updated in the supernet, groups are assigned by Expectation-Maximization clustering, and scoring functions are updated by reinforcement learning.
	

	\item We conduct extensive experiments on five popular benchmark data sets on link prediction and triplet classification tasks. 
	Experimental results demonstrate that ERAS can achieve state-of-the-art performance by designing relation-aware scoring functions.
	Especially, ERAS can consistently outperform literature at the relation type level of a given KG data.
	Besides, the search is much more efficient compared with AutoSF and the other popular automated methods.
\end{itemize}

\section{Related Works}
\label{sec:related}

\begin{table}[t]
	\centering
	\caption{Notations.}
	\label{table:notation}
	\vspace{-5px}
	\setlength\tabcolsep{4pt}
	\begin{tabular}{c|l}
		\hline
		\bf Symbols & \bf Meanings \\ \hline
		$E, R$     &       The entity set and relation set.        \\ \hline
		$S$     &  The KB  triples $\{(h,r,t)\}$, where $h,t\in E$ and $r\in R$.      \\ \hline
		$\langle\bm{h},\bm{r},\bm{t}\rangle$ &   The triple-dot product $\langle\bm{h},\bm{r},\bm{t}\rangle=\sum_i h_i\cdot r_i \cdot t_i$.          \\ \hline
		$\bm{\omega} = \{\bm{E}, \bm{R}\}$     &  The set of embeddings $E\!\in\!\mathbb{R}^{N_e\times d}$ and $R\!\in\!\mathbb{R}^{N_r\times d}$.  
		\\ \hline
		$f$     &   The scoring function, such as $f(\bm{h},\bm{r},\bm{t})$. \\ \hline	
		$\mathcal{M}$     &   The performance measurement. \\ \hline	
		$M,N$     &    The number of relation blocks and relation groups.           \\ \hline
		$\mathcal{O}$ & The operation set $\mathcal{O} \equiv \{\mathbf{0}, \pm\bm{r}_1, \cdots, \pm\bm{r}_M\}$. \\ \hline
		$\bm{A}$     &  The weight of architecture.  \\ \hline		
		$\bm{B}$     &   The weight of relation assignment.   \\ \hline		
	\end{tabular}
\end{table} 

\subsection{Neural Architecture Search (NAS)}
\label{ssec:automl}

\subsubsection{General Principles}\label{sec:rel:gen}
Generally, Neural Architecture Search (NAS)~\cite{automl_book,zoph2016neural,yao2018taking} is formed as a bi-level optimization problem~\cite{colson2007overview} where we need to update the neural architectures on the upper-level and train the model parameters in the lower-level.
Subsequently, three important aspects should be considered in NAS:
\begin{itemize}[leftmargin=*]
	\item \textit{Search space:} 
	it defines what network architectures in principle should be searched, e.g., Convolutional Neural Networks (CNNs)~\cite{lecun1998cnn} and Recurrent Neural Networks (RNNs)~\cite{rnn}.
	A well-defined search space should be expressive enough to enable powerful models to be searched. But it cannot be too large to search.
	
	\item \textit{Search algorithm:}
	it aims to efficiently search in the above space, e.g., bayesian optimization~\cite{bergstra2013making}, reinforcement learning~\cite{zoph2016neural}, evolution algorithm~\cite{xie2017genetic}.
	A search algorithm is required to perform an efficient search over the search space and be able to find architectures that achieve good performance.
	
	\item \textit{Evaluation mechanism:} it determines how to evaluate the searched architectures in the search strategy.
	Fast and accurate evaluation of candidate architectures can significantly boost the search efficiency.
	
\end{itemize}
Unfortunately, classic NAS~\cite{xie2017genetic,zoph2016neural} methods are computationally consuming since candidate architectures are trained by the \textit{stand-alone} way, i.e., many architectures are trained from scratch to convergence and are evaluated separately.

\subsubsection{One-shot Architecture Search (OAS)}
\label{ssec:nas}

More recently, One-shot Architecture Search (OAS) methods~\cite{pham2018efficient,liu2018darts,yao2019differentiable}, have been proposed to significantly reduce the search cost in NAS.
OAS first represents the whole search space by a \textit{supernet}~\cite{pham2018efficient}, which is formed by a directed acyclic graph (DAG), where the nodes are the operations in neural networks (e.g., $3\times 3$ conv in CNNs).
Every neural architecture in the space can be represented by a path in the DAG.
Then, instead of training independent model weights of each candidate architecture like the stand-alone approach, OAS keeps weights for the supernet and forces different architectures to share the same weights if they share the same edges in the DAG (i.e., \textit{parameter-sharing}~\cite{pham2018efficient,bender2018understanding}).
In this way, architectures can be searched by training the supernet once (i.e., the \textit{one-shot} manner), which makes NAS much faster.

Generally, OAS methods unify the search space with the form of DAG but differ in their way to search the optimal subgraph of DAG.
Sampling OAS (e.g., ENAS~\cite{pham2018efficient}) employs a controller to sample architectures and search an optimal subgraph of the whole DAG, which maximizes the expected reward of the subgraph on the validation set.
Instead of involving controllers, differentiable OAS (e.g., DARTS~\cite{liu2018darts} and NASP~\cite{yao2019differentiable}) relaxes the search space to be continuous so that the architectures can be optimized by gradient descent.
However, differentiable OAS may not able to derive an architecture that results in high evaluation performance when the evaluation metric is not differentiable.
In comparison, sampling OAS is more suitable for the non-differentiable scenario since it utilizes the policy gradient~\cite{williams1992simple} to optimize the controller.
	
\begin{table*}[t]
	\centering
		\caption{
			Hit@1 (in \%) results for existing scoring functions on the link prediction task.
		}
		\label{table:relationAware}
		\setlength\tabcolsep{8pt}
		\vspace{-5px}
		\begin{tabular}{c|c|cccc|cccc}
			\hline
			\multirow{2}{*}{SF Type} &\multirow{2}{*}{Methods} & \multicolumn{4}{c|}{\bf  Symmetric relations} & \multicolumn{4}{c}{\bf  Anti-symmetric relations} \\ \cline{3-10} 
			& &    WN18    &    WN18RR    &   FB15k    &    FB15k237   &    WN18      &      WN18RR    &     FB15k     &     FB15k237     \\ \hline
			\multirow{2}{*}{Non-universal} & TransE~\cite{bordes2013translating}	&   0.0     &    0.0    &    0.0  &   5.0    &   51.0    &   3.0  &  55.0   &    27.0   \\ \cline{3-10}
			& DistMult~\cite{yang2014embedding}	&    93.0    &   90.0     &   73.0    &   7.0    &      65.0     &       9.0   &      74.0    &     25.0     \\ \hline
			\multirow{5}{*}{Universal}& ConvE~\cite{dettmers2018convolutional}	&   93.0    &  93.0   & 42.0   &  1.0 &  94.0    & 6.0   &  61.0  &  25.0   \\ \cline{3-10}
			& TuckER~\cite{balazevic2019tucker}	&   94.0    &  93.0   &  67.0  &   2.0   &   95.0   &  12.0  &  73.0  &   22.0  \\ \cline{3-10}
			& ComplEx~\cite{trouillon2017knowledge}	&     94.0   &    94.0    &     88.0  &    2.0   &     95.0     &    11.0     &      80.0    &      23.0    \\ \cline{3-10}
			& SimplE~\cite{kazemi2018simple}	&    92.0    &     93.0   &    74.0   &    5.0   &   94.0 &  5.0    &   64.0       &  13.0      \\ \cline{3-10}
			& Analogy~\cite{liu2017analogical}	&  93.0    &   92.0    &   52.0    &   6.0    &    93.0      &     2.0     &     66.0     &   27.0     \\ \cline{3-10}
			& AutoSF~\cite{zhang2019autosf}	&  93.2  &  93.5 &  85.8  & 5.7 &    94.8   &  11.5   &   81.1   & 26.7  \\ \cline{1-10}
		\end{tabular}
\end{table*}

\subsection{AutoSF: Searching Task-aware Scoring Functions}
\label{ssec:asf}
Given a KG, it is very empirical to choose a suitable scoring function from the above manual methods. 
To better adapt to different KG tasks, AutoSF~\cite{zhang2019autosf} leverages the AutoML techniques to design and customize a proper scoring function on the given KG.

\subsubsection{Search Problem}
Motivated by the expressiveness guarantee and computational efficiency of BLMs, AutoSF proposes to partition embeddings $\bm{h},\bm{r},\bm{t} \in \mathbb{R}^d$ into $M$ splits (e.g., $\bm{h}=[\bm{h}_1;\cdots;\bm{h}_M]$ where $\bm{h}_i\in \mathbb{R}^{d/M}$), and represents scoring functions as:
\begin{align}
f(\bm{h}, \bm{r}, \bm{t})
= 
\sum\nolimits_{i=1}^M
\sum\nolimits_{j=1}^M
\langle\bm{h}_i,\bm{o},\bm{t}_j\rangle,
\label{eq:usf}
\end{align}
where 
$\bm{o} \in \mathcal{O}$ with
$\mathcal{O} 
\equiv 
\{\bm{0}, \pm\bm{r}_1, \cdots, \pm\bm{r}_M\}$. 
Note that
$\langle\bm{h}_i,\bm{o},\bm{t}_j\rangle$ computes the triple-dot product and is named as the \textit{multiplicative item}.
Then previous outstanding scoring functions~\cite{nickel2011three,yang2014embedding,liu2017analogical,trouillon2017knowledge,kazemi2018simple} can be unified in $f$ with different choices of $\bm{o}$~\cite{zhang2019autosf}.
This indicates that $f$ is general enough to represent good scoring functions which are designed manually.
In this way, AutoSF generalizes from human wisdom and allows the discovery of better scoring functions, which are not visited in the literature.
Subsequently,
the search problem is defined as:

\vspace{5px}
\begin{Definition}[AutoSF problem~\cite{zhang2019autosf}]
	\label{def:autoSF}\it
	Let $\bar{f}$ denote the desired scoring function
	and $\mathcal{F}_a$ denote the set of all possible scoring functions in AutoSF expressed by \eqref{eq:usf}. 
	Then the scoring function search problem is defined as follows:
	\begin{align*}
	\bar{f}
	& = 
	\arg\max\nolimits_{f \in \mathcal{F}_a} 
	\mathcal{M}_{\text{val}}
	\left (
	f, \bar{\bm{\omega}}; S_{\text{val}}
	\right)
	\\
	\text{ s.t. }
	\bar{\bm{\omega}}
	& = 
	\arg\max\nolimits_{\bm{\omega}}
	\mathcal{M}_{\text{tra}}
	\left (
	f, \bm{\omega}; S_{\text{tra}}
	\right),
	\end{align*}
	where
	$\mathcal{M}_{\text{tra}}$
	and
	$\mathcal{M}_{\text{val}}$
	measure the performance of scoring function $f$ and KG embeddings $\bm{\omega}$ on corresponding KG data $S$ (e.g., training set $S_{\text{tra}}$ and validation set $S_{\text{val}}$), respectively.
\end{Definition}

\vspace{5px}

Given the embeddings $\bar{\bm{\omega}}$ learned on the training data $S_{\text{tra}}$, AutoSF aims to search for a better scoring function $f$ which leads to higher performance on the validation set $S_{\text{val}}$.
Hence, AutoSF can find task-aware scoring functions that can achieve impressive performance on different KG tasks.
However, it is non-trivial to efficiently search a proper scoring function from the AutoSF's search space due to its size  $O((2M+1)^{M^2})$.

\subsubsection{Search Algorithm}
\label{sssec:autosfSearch}
Since a large number of possible scoring functions exist in the unified scoring function search space, 
AutoSF then proposes a progressive greedy search algorithm to find a proper scoring function according to an inductive rule:
\begin{equation}
\label{eq:inductive}
f^b = f^{b-1} + 
\langle\bm{h}_i,\bm{o},\bm{t}_j\rangle
,
\end{equation}
where $b$ is the burget of non-zero multiplicative terms (i.e., $\bm{o}\neq \mathbf{0}$) in \eqref{eq:usf}. 
The intuition behind \eqref{eq:inductive} is to gradually add nonzero multiplicative terms $\langle\bm{h}_i,\bm{o},\bm{t}_j\rangle$ to achieve the final desired $f$.
Each greedy step in the search algorithm mainly contains two parts: 
sampling scoring functions and evaluate the sampled scoring functions.
We summarize the search algorithm in Algorithm~\ref{alg:autosf}.

\begin{algorithm}[ht]
	\caption{AutoSF: Progressive Greedy Search of Task-aware Scoring Functions}
	\label{alg:autosf}
	\begin{algorithmic}[1]
		\renewcommand{\algorithmicrequire}{\textbf{Input:}}
		\renewcommand{\algorithmicensure}{\textbf{Output:}}
	\REQUIRE $B$: number of nonzero blocks.
		\FOR{$b$ in $4,\cdots,B$}
		\STATE Randomly select $N$ scoring functions $\{f^{b-1}\}$;
		\STATE Sample $N_1$ scoring functions $\{f^{b}\}$ by adding relation blocks to $\{f^{b-1}\}$ as $f^b = f^{b-1} +\langle\bm{h}_i,\bm{o},\bm{t}_j\rangle$;
		\STATE Select top-$K$ $\{f^b\}$ based on the \textit{Predictor}.
		\STATE Train the top-$K$ $\{f^b\}$ to convergence separately and update \textit{Predictor} with the evaluated performance.
		\ENDFOR
		\RETURN Scoring function $f^B$ with highest performance.
	\end{algorithmic}
\end{algorithm}

However, there are several issues in AutoSF.
First, the evaluation mechanism in AutoSF is inefficient.
In every greedy step, 
AutoSF trains all candidate scoring functions under budget $b$ to convergence for performance evaluation, i.e., step 5 of Algorithm~\ref{alg:autosf}.
Then well-trained embeddings of all scoring functions will be discarded in the next greedy search. 
It wastes a lot of effort to train KG embeddings for performance evaluation.
Moreover, within the budget $b$, the predictor in AutoSF search algorithm can only leverage the prior experience that is smaller than the budget $b$, i.e., step 4 of Algorithm~\ref{alg:autosf}, which may also bring unnecessary KG embeddings training due to inaccurate prediction.
Second, AutoSF pursuits a universal scoring function over a given KG data. 
A universal scoring function that can learn certain relationships does not necessarily mean that this scoring function can perform well on them~\cite{meilicke2018fine,rossi2020knowledge}.
This will be further discussed in Section~\ref{ssec:motivation}.

\section{Problem Definition}
\label{sec:probdef}

In this section, to further illustrate the motivation of relation-aware scoring functions, we first discuss the performance of existing scoring functions at the relation pattern level.
Then, we formulate a relation-aware scoring function search problem.

\subsection{Motivation of Relation-aware Scoring Functions}
\label{ssec:motivation}

As introduced in Section~\ref{sec:intro}, traditional scoring functions and AutoSF try to design universal scoring functions to cover as many as relation patterns as possible. 
However, being expressive does not mean achieving good performance as relations exhibit different patterns~\cite{meilicke2018fine,rossi2020knowledge}.
We summarize the experimental results reported by~ Figure 14 and 15 from \cite{rossi2020knowledge} in Table~\ref{table:relationAware}, which demonstrate the performance (the higher the better) of popular scoring functions on symmetric and anti-symmetric relations (e.g., Table~\ref{tab:relationPattern}).

\begin{table}[ht]
	\centering
	\caption{Exemplar relations corresponding to relation patterns in the benchmark data sets
		(see Section~\ref{sec:exp:data} for details).}
	\label{tab:relationPattern}
	\vspace{-5px}
	\begin{tabular}{c|c|c}
		\hline
		\bf Relation Patterns & \bf  WN18/WN18RR& \bf FB15k/FB15k237\\ \hline
		Symmetric &   \textit{similar\_to, synset\_of} & \textit{spouse\_of}\\ \hline
		Anti-symmetric & \textit{hypernym, hyponym}  &  \textit{child\_of} \\ \hline
	\end{tabular}
\end{table}

	From Table~\ref{table:relationAware}, it is worth noting that 
	universal scoring functions may perform even worse than non-universal scoring functions at the relation pattern level.
	First, TransE performs badly on symmetric relations on all benchmark data sets~\cite{bordes2013translating,toutanova2015observed,dettmers2018convolutional} since it cannot handle the symmetric relations.
	But it achieves better performance on symmetric relations of FB15k237~\cite{toutanova2015observed} than several universal scoring functions, such as ConvE, TuckER, ComplEx.
	
	Second, DistMult only covers symmetric relations. Therefore, DistMult achieves good performance on symmetric relations, while it performs unsatisfactorily on anti-symmetric relations.
	However, as reported in Table~\ref{table:relationAware},
	there are several cases that universal scoring functions perform worse than DistMult on anti-symmetric relations:
	\begin{itemize}[leftmargin=12px]
		\item[1.] 
		ConvE, SimplE, and Analogy in WN18RR~\cite{dettmers2018convolutional}.
		
		\item[2.] 
		ConvE, SimplE, and Analogy in FB15K~\cite{bordes2013translating}.
		
		\item[3.] 
		TuckER, ComplEx, SimplE, and AutoSF in FB15k237~\cite{toutanova2015observed}.
	\end{itemize}

In summary, universal scoring functions may achieve unsatisfactory performance on specific relation patterns of certain KG.
Such observation motivates us to design relation-aware scoring functions.

\subsection{Problem Formulation}
\label{ssec:prodef}

Inspired by the observation in Section~\ref{ssec:motivation} and the
task-aware method AutoSF, we here propose to search relation-aware scoring functions for different relation patterns over any given KG data.

Recall that AutoSF targets to find a scoring function $f$ that can achieve high $\mathcal{M}(f,\bm{\omega};S)$ for given triplets $S$.
But in relation-aware search, it is also important to assign relations to appropriate groups to better cover relation patterns.
Let $\bm{B} \in \{ 0, 1 \}^{N_r\times N}$ record the relation assignment, where $B_{rn}=1$ if the $r\in R$ is assigned to $n$-th group otherwise $B_{rn}=0$,
and $f_n$ denote the scoring function for relations in $n$-th group.
Then, relation-aware search aims to find a set of scoring functions $\{f_n\}$ and relation assignments $\bm{B}$ that can achieve high $\mathcal{M}(\{f_n\},\bm{B},\bm{\omega};S)$. Formally, the problem in this paper is defined as:

\vspace{5px}
\begin{Definition}[ERAS problem]
	\label{def:problem} \it
	Let $N$ denote the total number of relation groups, and $\mathcal{F}_e$ denote the set of all possible relation-aware scoring functions.	
	Then the relation-aware scoring function search problem is defined as:
	\begin{align}
	\label{eq:objective}
	& \{ \bar{f}_n \}_{n=1}^N
	= \arg\max\nolimits_{f_n \in  \mathcal{F}_e }
	\
	\mathcal{M}_{\text{val}} \left(\{f_n\}_{n=1}^N,\bar{\bm{B}},\bar{\bm{\omega}};S_{\text{val}}\right),
	\\
	& \text{ s.t. } 
	\begin{cases}
	\bar{\bm{\omega}}
	= 
	\arg\max\nolimits_{\bm{\omega}}\;
	\mathcal{M}_{\text{tra}} \left(\{f_n\}_{n=1}^N,\bar{\bm{B}},\bm{\omega};S_{\text{tra}}\right)
	\\
	\bar{\bm{B}}
	= \arg\min\nolimits_{\bm{B}} \;
	\mathcal{L}_{B}
	\left(
	\bm{B},
	\bm{\omega}
	\right)
	\end{cases}
	\!\!\!\!\!\!,
	\label{eq:objective2}
	\end{align} 
	where $\mathcal{L}_B$ is the loss of relation assignments,
	and
	$\mathcal{M}_{\text{tra}}$,
	$\mathcal{M}_{\text{val}}$
	measures the performance on 
	the training set $S_{\text{tra}}$ and validation set $S_{\text{val}}$, respectively.
\end{Definition}
\vspace{5px}

Generally, the relation-aware scoring function search problem is based on the bi-level optimization problem in Definition~\ref{def:problem}. 
Compared with the single-level objective, bi-level optimization allows the model to optimize parameters in different ways, which is more suitable for deriving scoring functions, embeddings, and relation assignments.
This definition looks similar to AutoSF in Definition~\ref{def:autoSF}, but it is quite different in essence. 
First, we need to assign relations to proper relation groups in the lower-level objective  \eqref{eq:objective2}.
Second, there are multiple targeted scoring functions $\{f_n\}_{n=1}^N$ for handling $N$ relation groups.
Therefore, the search space $\mathcal{F}_e$ for ERAS with size $O((2M+1)^{NM^2})$  is much larger than that for AutoSF with size $O((2M+1)^{M^2})$.
The larger search space requires ERAS to have a more efficient search algorithm.

\section{Search Algorithm}
\label{sec:algorithm}

Here, we propose a new algorithm to solve the search problem in Definition~\ref{def:problem}.
We can see that there are three types of parameters need to be simultaneously optimized in \eqref{eq:objective}, i.e.,
\begin{itemize}[leftmargin=*]
	\item \textit{Group Assignments}:
	The relation assignment mechanism should be flexible enough to update $\bm{B}$ during the whole search procedure.
	
	\item \textit{Architectures}:
	We pursue $\{f_n\}_{n=1}^N$ with high performance. 
	But it is difficult to maximize \eqref{eq:objective} because performance evaluation in KG embedding is usually non-differentiable.
	
	\item \textit{Embeddings}:
	Training KG embeddings $\bm{\omega}$ for evaluating candidate scoring functions consume a lot of computation overhead in AutoSF.
	It is essential to tackle this issue since ERAS has a much larger search space than AutoSF.
\end{itemize}

Due to these challenges, neither AutoSF nor other existing NAS algorithms can be applied (see discussions in Section~\ref{sec:comp:autosf} and \ref{ssec:comp:oas}).
Thus, we propose to deal with the above challenges using alternative minimization.
Specifically, we incorporate three key components in search algorithm: \textit{Expectation-Maximization (EM) clustering} for updating $\bm{B}$, \textit{policy gradient} for searching $\{f_n\}_{n=1}^N$, and \textit{embedding sharing} for updating $\bm{\omega}$.
Details are in the sequel.

\begin{figure*}[t]
		\centering
		\subfigure[Modelling the generation of $f$ as a multi-step decision process.] 
		{\includegraphics[width=0.85\columnwidth]{Figs/sampling/rnn2_.png}}
		\quad
		\subfigure[Architecture of supernet.] 
		{\includegraphics[width=0.45\columnwidth]{Figs/sampling/supernet2_.png}}
		\subfigure[Illustration to embedding sharing.] 
		{\includegraphics[width=0.55\columnwidth]{Figs/sampling/embedShare_.png}}
		\caption{
				Set $M,N=2$ in all examples.
				(a) An example of recurrently generating the relation-aware scoring functions 
				$\{f_1,f_2\}$: $f_{1}(\bm{h},\bm{r},\bm{t}) = \left\langle \bm{h}_1, \bm{r}_1, \bm{t}_1 \right\rangle 
				+ \left\langle \bm{h}_2, \bm{r}_2, \bm{t}_2 \right\rangle $ 
				and $f_{2}(\bm{h},\bm{r},\bm{t}) = - \left\langle \bm{h}_1, \bm{r}_1, \bm{t}_2 \right\rangle  
				-\left\langle \bm{h}_2, \bm{r}_2, \bm{t}_1 \right\rangle $;
				(b) The illustration to supernet in the form of bipartite graph and architecture weight $\bm{A}$;
				(c) The example of sharing embeddings between two sampled relation-aware scoring functions.
		}
		\label{figs:framework}
	\vspace{-5px}
\end{figure*}

\subsection{Update Group Assignments by EM Clustering}
\label{ssec:clustering}
In this part, we illustrate how to assign relations to proper groups in the search procedure.
Intuitively, relations with similar semantic meanings should be grouped.
Since the KG embeddings are designed to encode the semantic meanings \cite{bordes2013translating,wang2017knowledge}, we propose to assign relations based on the given KG embeddings $\bm{\omega}$, i.e., minimizing $\mathcal{L}_B(\bm{B},\bm{\omega})$ in Definition~\ref{def:problem}.
Specifically, given a set of relations $R$ and $N$ groups, let $\bm{c}_n$ denote the vector representation of the $n$-th group (i.e., $\bm{C}$ for all groups), and $B_{rn}$ define the degree of membership between the relation $r$ with $n$-th group.
Then, the objective $\mathcal{L}_{B}$ for relation clustering in Definition~\ref{def:problem} is defined as:
\begin{equation}
\label{eq:SSE}
\min\nolimits_{\bm{B}}
\mathcal{L}_{B}(\bm{B},\bm{\omega})
\! = \! \min\nolimits_{\bm{B}, \bm{C}} 
\sum\nolimits_{r}
\sum\nolimits_{n} B_{rn}
\!
\left \Vert 
\bm{r}
\!-\!
\mathbf{c}_n
\right\Vert^2
,
\end{equation}
which can be solved by the Expectation-Maximization (EM) algorithm~\cite{dempster1977maximum}. 

\subsection{Updating Architectures by Reinforcement Learning}
\label{ssec:stochastic}

Our target in \eqref{eq:objective} is to derive an optimal relation-aware scoring function $\{\bar{f}_n\}_{n=1}^N$, which maximizes the evaluation performance on the given KG data.
It is natural that we use Mean Reciprocal Ranking (MRR) on the validation data as the evaluation measurement $\mathcal{M}_{\text{val}}$.
However, MRR is non-differentiable, which indicates that directly optimizing \eqref{eq:objective} by gradient descent (e.g., differentiable OAS \cite{liu2018darts,yao2019differentiable}) is not suitable (see discussions in Section~\ref{sssec:ablationEvaluate}).
	
\subsubsection{Reinforcement Learning Formulation}
To handle the non-differentiable MRR, we first formulate 
the  search problem of $\{f_n\}$ as a multi-step decision problem. 
Then we adopt reinforcement learning to solve \eqref{eq:objective}.

Recall that each $f_n$ is a summation of multiplication terms $\langle\bm{h}_i,\bm{o},\bm{t}_j\rangle$ in \eqref{eq:usf}.
We can sequentially determine which operation $\bm{o}\in \mathcal{O}$ is selected for the $(i,j)$-th multiplicative item in $f_n$.
Let $\emph{v}$ denote the index of all multiplicative items in $\{f_n\}$ and $\alpha_{\emph{v}}$ denote the operation selected for the $\emph{v}$-th multiplicative item.
Then, as shown in Figure~\ref{figs:framework}~(a), the search process of $\{f_n\}$ can be viewed as a multi-step decision problem:
a list of tokens $\{\alpha_{\emph{v}}\}_{\emph{v}=1}^{V}$ ($V=NM^2$) that needs to be predicted.
It is intuitive to adopt reinforcement learning to solve this problem.
Let $\bm{A} \in \{0,1\}^{NM^2\times(2M+1)}$, where $A_{\emph{v}k}=1$ if $\alpha_\emph{v}$ chooses $k$-th operation $\bm{o}_k$ in $\mathcal{O}$ otherwise $A_{\emph{v}k}=0$.
Then \eqref{eq:objective} can be reformulated as:
\begin{align}
\max\nolimits_{\theta}
\mathcal{J}(\theta) 
\equiv 
\mathbb{E}_{\bm{A} \sim \pi(\bm{A}; \theta)}[Q \left(\bm{A},\bm{B},\bm{\omega};S_{\text{val}}\right)],
\label{eq:updateA}
\end{align}
where  $\pi(\bm{A}; \theta)$ is a policy parameterized by $\theta$ for generating $\bm{A}$, and $Q \left(\bm{A}, \bm{B}, \bm{\omega}; S_{\text{val}}\right)$ measures MRR performance as:
\begin{align*}
Q \left(\bm{A}, \bm{B}, \bm{\omega}; S_{\text{val}}\right)
=
\sum\nolimits_{n}
\sum\nolimits_{(h,r,t)\in S_{\text{val}}}
B_{rn}\cdot
q(f_n(\bm{h},\bm{r},\bm{t})).
\end{align*}
Note that $q(\cdot)$ measures MRR of a triplet and $f_{n}$ is the $n$-th scoring function based on $\bm{A}$.
Then, the gradient of $\theta$ in \eqref{eq:updateA} can be computed by REINFORCE gradient~\cite{williams1992simple} as 
\begin{align}
& \nabla_{\theta}\mathcal{J}(\theta)
=
\mathbb{E}_{\bm{A}\sim \pi(\bm{A}; \theta)}
\left[ 
\sum\nolimits_{\emph{v}=1}^{V}  
\nabla_{\theta}
\log{P(\alpha_\emph{v}|\alpha_{(\emph{v}-1):1};\theta)}
Q
\right]
\notag
\\
\label{eq:controllerUpdate2}
& \approx
\frac{1}{U} \!
\sum\nolimits_{u=1}^{U}
\sum\nolimits_{\emph{v}=1}^{V}
\nabla_{\theta}
\log{P(\alpha_\emph{v}|\alpha_{(\emph{v}-1):1};\theta)}
(Q_u \! - \! b),
\end{align}
where $Q_u$ denotes $Q\left(\bm{A}^u,\bm{B},\bm{\omega};S_{\text{val}}\right)$, 
$\bm{A}^u$ is 
$u$-th sampled architecture from $\pi(\bm{A};\theta)$,
$b$ is a moving average baseline to reduce variance, 
and $U$ denote the number of sampled scoring functions.
We can see that solving \eqref{eq:updateA} has been converted to optimize $\theta$ in  \eqref{eq:controllerUpdate2}, and whether $Q$ is differentiable does not affect the gradient computation w.r.t $\theta$.

Inspired by \cite{zoph2016neural,pham2018efficient}, we also adopt a Long Short-term Memory (LSTM)~\cite{hochreiter1997long} to parameterize $\theta$ for learning the policy $\pi(\bm{A};\theta)$.
Specifically, the controller samples decisions in an autogressive way: the decided operation $\alpha_\emph{v}$ in the previous multiplicative item is carried out and then fed into the next to predict $\alpha_{\emph{v}+1}$ (see Figure~\ref{figs:framework} (a)).
Finally, \eqref{eq:controllerUpdate2} is used to update the LSTM policy network $\theta$.

\subsubsection{Encoding Prior of Architectures}
To fully evaluate the search space, we expect that every relation segmentation $\bm{r}_i\in\mathcal{O}\backslash\mathbf{0}$ must be selected at least once in the searched scoring functions, named as the \textit{exploitative constraint}. 
Thus, we constrain $\bm{A}$ with this exploitative constraint.
If the sampled $\bm{A}$ does not satisfy it, we will directly set the reward $Q$ to 0.

\subsection{Update Embedding in Shared Supernet}
\label{ssec:parametersharing}

AutoSF follows the classic NAS way to evaluate the stand-alone performance of candidate scoring functions, which requires separately training KG embeddings hundreds of times.
As discussed in Section~\ref{ssec:nas}, OAS methods propose a more efficient evaluation mechanism by forcing all candidates sharing parameters.
Inspired by OAS works, we design a simple but effective \textit{supernet} that enables candidate scoring functions sharing the KG embeddings for accelerating search.


\begin{table*}[ht]
	\caption{Comparison of AutoSF (Algorithm~\ref{alg:autosf})
		and ERAS (Algorithm~\ref{alg:ERAS})
		in terms of NAS principles (Section~\ref{sec:rel:gen}).}
	\label{tab:techcomp}
	\centering
	\vspace{-5px}
	\begin{tabular}{c | c  c | C{120px} | c}
		\hline
		&   \multicolumn{2}{c|}{\textbf{space}}   & \multirow{2}{*}{\textbf{algorithm}}                            & \multirow{2}{*}{\textbf{evaluation}} \\
		& size             & property             &                                                                &                                      \\ \hline
		AutoSF & $O( (2M+1)^{M^2}) $  & task-aware           & progressive greedy search                                      & by stand-alone                       \\ \hline
		ERAS  & $O( (2M+1)^{NM^2}) $ & task-/relation-aware & alternative minimization (EM cluster + reinforcement learning) & in embedding shared supernet         \\ \hline
	\end{tabular}
\end{table*}

\subsubsection{Design of Supernet}
To enable fast evaluation, we propose a supernet view of the relation-aware scoring function search space. 
Specifically, the $f_n$ is represented as:
\begin{align}
f_n (\bm{h}, \bm{r}, \bm{t})
=
\sum\nolimits_i \sum\nolimits_j
\sum\nolimits_k
A_{\emph{v}k} 
\cdot
\langle\bm{h}_i,\bm{o}_{k},\bm{t}_j\rangle.
\label{eq:sup}
\end{align}
Recall that $\emph{v}$ is the index of multiplicative items in $\{f_n\}$ and $k$ is the index of operations in $\mathcal{O}$.
Then, as shown in Figure~\ref{figs:framework}~(b), we can take $\bm{A}$ as the adjacency matrix of a bipartite graph (i.e., supernet) from above \eqref{eq:sup}, where multiplicative items and operations are nodes, and $A_{\emph{v}k}$ records the edge weight between multiplicative items and operations.
Based on the supernet design, Figure~\ref{figs:framework}~(c) illustrates that any relation-aware $\{f_n\}$ can be realized by taking subgraphs of the supernet.
Then ERAS forces all subgraphs to share embeddings, thereby 
different scoring functions can be evaluated based on the same KG embedding.
It enables us to evaluate candidate scoring functions faster by avoiding repetitive embedding training.

\subsubsection{Update Embeddings}
Given the fixed controller's policy $\pi(\bm{A};\theta)$ and relation assignment $\bm{B}$, we propose to solve $\mathcal{M}_{\text{tra}}$ in objective \eqref{eq:objective2} by minimizing the expected loss $\mathcal{L}$ on the training data, such as $\mathbb{E}_{\bm{A}\sim \pi(\bm{A}; \theta)}[\mathcal{L} \left(\bm{A},\bm{B},\bm{\omega};S_{\text{tra}}\right)]$.
Then stochastic gradient descent (SGD) can be performed to optimize $\bm{\omega}$.
We approximate the gradient $\nabla_{\bm{\omega}}$ according to:
\begin{equation}
\label{eq:embeddingUpdate}
\nabla_{\bm{\omega}} \mathbb{E}_{\bm{A}\sim \pi(\bm{A}; \theta)}[\mathcal{L}
] 
\approx
\frac{1}{U} 
\sum\nolimits_{u=1}^{U} 
\nabla_{\bm{\omega}} \mathcal{L} \left(\bm{A}^{u},\bm{B},\bm{\omega};S_{\text{tra}}\right), 
\end{equation}
where $U$ is the number sampled scoring functions, and
\begin{align*}
\mathcal{L}\left(\bm{A}^u,\bm{B},\bm{\omega};S_{\text{tra}}\right) 
= \sum\nolimits_n  \sum\nolimits_{(h,r,t) \in S_{\text{tra}}} B_{rn} \cdot \ell(f_{n} (\bm{h},\bm{r},\bm{t})).
\end{align*}
Note that $\ell(\cdot)$ is the 
multiclass log-loss~\cite{lacroix2018canonical} and
$f_{n}$ is the $n$-th scoring function based on sampled $\bm{A}^u$.
Hence,
\eqref{eq:embeddingUpdate} can be represented as:
\begin{align*}
\nabla_{\bm{\omega}} \mathbb{E}_{\bm{A}\sim \pi(\bm{A}; \theta)}[\mathcal{L}
] 
\!\approx\!
\frac{1}{U}
\sum\nolimits_{u=1}^{U}
\sum_n
\!\!
\sum_{(h,r,t) \in S_{\text{tra}}} 
\!\!\!\!\!\!
\nabla_{\bm{\omega}}
B_{rn} 
\!\cdot\! 
\ell(f_{n} (\bm{h},\bm{r},\bm{t})).
\end{align*}

\subsubsection{Performance Evaluation}
Recall that $Q_u$ denotes the reward of the $u$-th sampled scoring function in \eqref{eq:controllerUpdate2}.
In the supernet, every sampled scoring function can be regarded as a subgraph of the supernet where some edges are activated.
Hence, we can evaluate $Q_u$ based on the sampled scoring functions by activating the subgraph on the supernet.

\subsection{Complete Algorithm}

The proposed algorithm ERAS is summarized in Algorithm~\ref{alg:ERAS}, where KG embedding $\bm{\omega}$, relation assignments $\bm{B}$ and architectures $\bm{A}$ are alternatively updated in every epoch.
To improve the efficiency of scoring function search, we represent the search space as a supernet and propose to share KG embeddings across different scoring functions in step~3 (see Section~\ref{ssec:parametersharing}).
Thus, ERAS is capable of avoiding wasting a lot of computation on training embeddings from scratch.
To enable relation-aware scoring function search, we introduce EM clustering in step~4 to dynamically assign relations $\bm{B}$ based on the learned embeddings (see Section~\ref{ssec:clustering}).
To handle the non-differentiable measurement of scoring functions, we use reinforcement learning and perform policy gradient in step~5-6 (see Section~\ref{ssec:stochastic}). 
After searching, we derive several sampled scoring functions with the well-trained controller and compute its reward on a mini-batch of the validation data in step~9-11. 
Finally, we take the scoring functions with the highest reward and re-train it from scratch.

\begin{algorithm}[ht]
	\caption{ERAS: Efficient Relation-aware Scoring Functions Search.}
	\label{alg:ERAS}
	\begin{algorithmic}[1]
		\renewcommand{\algorithmicrequire}{\textbf{Input:}}
		\renewcommand{\algorithmicensure}{\textbf{Output:}}
		\REQUIRE Initialize embeddings $\bm{\omega}$, relation groups $\bm{B}$, and controller's parameter $\theta$.
		\\ // search relation- and task-aware scoring functions
		\WHILE{not converge}
		\STATE Sample a set of scoring functions from $\pi(\bm{A};\theta)$;
		\STATE Update \textit{shared embeddings} $\bm{\omega}$ with \eqref{eq:embeddingUpdate};	
		\STATE Update \textit{relation assignments} $\bm{B}$ according to  \eqref{eq:SSE};
		\STATE Sample a mini-batch data $B_{\text{val}}$ from validation data $S_{\text{val}}$;
		\STATE Update \textit{architectures} policy $\pi(\bm{A}; \theta)$ with \eqref{eq:controllerUpdate2};
		\ENDWHILE
		\\ // Derive the final scoring function $\bar{\bm{A}}$
		\STATE  ERAS samples $K$ scoring functions $\mathcal{A}^{K}$ from $\pi(\bm{A};\theta)$;
		\FOR{$\bm{A}\in \mathcal{A}^{K}$}
		\STATE Compute $Q \left(\bm{A},\bm{B},\bm{\omega};S_{\text{val}}\right)$;
		\ENDFOR
		\RETURN The scoring function $\bar{\bm{A}}$ with highest reward on validation set,  
		and train it from scratch to convergence.
	\end{algorithmic}
\end{algorithm}

\subsubsection{Comparison with AutoSF}\label{sec:comp:autosf}
To compare ERAS with the pioneering work AutoSF, we summarize them from the perspective of NAS principles in Table~\ref{tab:techcomp}.
First, to capture the inherent properties of relation patterns in KGs, 
ERAS targets to solve the relation-aware scoring function search problem in this paper.
This leads to the larger search space in ERAS than that in AutoSF.
Second, ERAS proposes an alternative minimization way to solve the non-differentiable problem of $\mathcal M_{\text{val}}$.
Finally, AutoSF evaluates every scoring function based on their stand-alone performance, which requires repeatedly training KG embeddings hundreds of times.
On the contrary, ERAS shares KG embeddings across different scoring functions, which extremely reduces time expense for performance evaluation of candidate scoring functions.

\subsubsection{Comparison with Existing OAS Algorithms}\label{ssec:comp:oas}
Inspired by parameter sharing in OAS works, we propose to share KG embedding in the supernet, thereby ERAS avoids repeatedly training KG embedding hundreds of times.
However, there exist some concerns about parameter-sharing in OAS \cite{li2019random,bender2018understanding}.
Specifically, while parameter sharing enables an alternative way to train all architectures in the search space in a cheaper way, it can lead to a biased evaluation problem: the evaluation of candidate architectures in the search phase (i.e., $\mathcal{M}_{\text{val}}$) is inconsistent with the stand-alone training phase.
Especially, the correlation between one-shot and stand-alone performance will probably be unstable as the supernet goes deep and complex.

Therefore, to make parameter sharing work for our problem,
we first construct a search space, which can be represented by a supernet in \eqref{eq:sup}.
Then, to avoid the supernet being deep and complex, we design a shallow and simple supernet with the form of a bipartite graph, which differs from the complex DAG supernet in classic OAS works.
We demonstrate that the ERAS's supernet design does not suffer from a biased evaluation problem in Section~\ref{sssec:ablationEvaluate}.
In summary, we leverage the domain-knowledge of KG embedding to make embedding sharing works.

\begin{table*}[t]
	\caption{
		Comparison of the best scoring functions identified by ERAS and the state-of-the-arts on the link prediction task.
		The bold numbers mean the best performance and the underline ones mean the second best.
		$[\clubsuit]$: results are taken from \cite{zhang2019autosf};
		$[\dag]$: from \cite{sun2019rotate};
		$[\ddag]$: from \cite{socher2013reasoning};
		$\S$: from \cite{balavzevic2019hypernetwork};
		$[\diamondsuit]$: from \cite{dettmers2018convolutional};
		$[*]$: from \cite{balazevic2019tucker};
		$[+]$: from \cite{xue2018expanding};
		$[\spadesuit]$: from \cite{zhang2019quaternion}.
	}
	\label{table:linkPrediction}
	\setlength\tabcolsep{3pt}
	\vspace{-3px}
	\centering
	\begin{tabular}{c|c|ccc|ccc|ccc|ccc|ccc}
		\hline
		\multirow{2}{*}{type} & \multirow{2}{*}{model} &      \multicolumn{3}{c|}{\bf WN18}       &     \multicolumn{3}{c|}{\bf WN18RR}      &      \multicolumn{3}{c|}{\bf FB15k}      &    \multicolumn{3}{c|}{\bf FB15k237}     &     \multicolumn{3}{c}{\bf YAGO3-10}     \\ \cline{3-17}
		&                        &        MRR       &      Hit@1  &      Hit@10      &        MRR    &      Hit@1    &      Hit@10      &        MRR    &      Hit@1    &      Hit@10      &        MRR     &      Hit@1   &      Hit@10      &        MRR  &      Hit@1      &      Hit@10      \\ \hline
		&         TransE$^\clubsuit$   &       0.500  &   -  &       94.1       &      0.178       & -   &    45.1       &       0.495   & -    &       77.4       &       0.256       &   -  &  41.9       &         -     & -   &        -         \\
		TDMs         &         TransH$^\clubsuit$        &       0.521     & - &       94.5       &       0.186    &   &       45.1       &       0.452      & - &       76.6       &       0.233      & - &       40.1       &         -    & -    &        -         \\
		
		&         RotatE$^\dag$        &       0.949    &  94.4 &       95.9       &       0.476  &  42.8   &       57.1       &       0.797      & 74.6 &       88.4       &     0.338      & 24.1 &        53.3      &         -     & -    &        -         \\ \hline
		&          NTN$\ddag$        &       0.53    &  -   &       66.1       &         -         &  -   &    -         &       0.25      & -   &       41.4       &         -         &  -   &    -         &         -    &      &        -         \\
		NNMs         &         ConvE$^\diamondsuit$        &       0.942       & 93.5    &   95.5       &       0.460      & 39.0 &       48.0       &       0.745    & 67.0   &       87.3       &       0.316      & 23.9 &       49.1       &       0.520      &  45.0 &       66.0       \\			
		&         HypER$^\S$        &       0.951       &    94.7  & 95.8       &       0.465      &  43.6 &       52.2       &       0.790    & 73.4   &       88.5       &       0.341       & 25.2 &       52.0       &       0.533      & 45.5  &       67.8       \\ \hline
		
		\multirow{7}{*}{TBMs}  &         TuckER  $*$       &     \bf 0.953 & 
		\underline{94.9}    &       95.8       &       0.470   &  44.3  &       52.6       &       0.795   &   74.1 &       89.2       &       0.358    &  26.6 &       54.4       &         -      &-   &        -         \\
		&         HolEX$^+$        &       0.938      & 93.0 &       94.9       &         -        & - &        -         &       0.800    & 75.0   &       88.6       &         -        & - &        -         &         -      &  -  &        -         \\
		
		&         QuatE$^\spadesuit$         &       0.950     & 94.5  &       95.9       &       0.488     & 43.8  &     \bf 58.2     &       0.782     & 71.1  &       90.0       &       0.348    &  24.8  &       55.0      &         -     & -    &        -         \\
		
		&        DistMult$^\clubsuit$       &       0.821       & 71.7  &    95.2       &       0.443    & 40.4  &       50.7       &       0.817    & 77.7  &       89.5       &       0.349      & 25.7 &       53.7       &       0.552     & 47.6  &       69.4       \\
		
		&        ComplEx$^\clubsuit$       &       0.951       &  94.5  &   95.7       &       0.471      &  43.0 &       55.1       &       0.831    &  79.6 &       90.5       &       0.347    &  25.4 &       54.1       &       0.566   &  49.1  &       70.9       \\
		
		&        Analogy$^\clubsuit$        &       0.950       &  94.6 &   95.7       &       0.472   &  43.3  &       55.8       &       0.829   &  79.3  &       90.5       &       0.348     & 25.6  &       54.7       &       0.565    & 49.0  &       71.3       \\
		&         SimplE$^\clubsuit$      &       0.950       & 94.5 &      95.9       &       0.468     & 42.9  &       55.2       &       0.830     & 79.8  &       90.3       &       0.350     & 26.0  &       54.4       &       0.565    &  49.1  &       71.0       \\ \hline
		Rule-based &         AnyBURL    &    0.950  & 94.6 &  95.9        &   0.480   & 44.6  &  55.5    &    0.830   &  80.8 &  87.6   &     0.310 & 23.3  &   48.6      &    0.540   &  47.7  &   67.3      \\ \hline
		\multirow{2}{*}{AutoML}       &         AutoSF$^\clubsuit$       & \underline{0.952} & 94.7 &\underline{96.1} & \underline{0.490} & \underline{45.1} &       56.7       & \underline{0.853} &\underline{82.1}  & 91.0 & 0.360 & \underline{26.7} & \underline{55.2} & \underline{0.571} & 50.1 & \underline{71.5} \\
		& $\text{ERAS}^{N=1}$  &  0.951 & 94.7   & 96.0  &   \underline{0.490} & 45.0 &  \underline{56.8} & \underline{0.853} &  82.0  & \underline{91.2}  & \underline{0.361}  &26.6  &  \underline{55.2}  & 0.570 & \underline{50.2} & \underline{71.5} \\
		&  ERAS  &     \bf 0.953     &  \bf 95.0 & \bf 96.2     &     \bf 0.492   & \bf 45.2 & \underline{56.8} &    \bf{0.855} & \bf 82.3   &    \bf{91.4}     &     \bf 0.365   &  \bf26.8 &     \bf 55.5     &     \bf 0.577    & \bf50.3 &    \bf   71.7    \\ \hline
	\end{tabular}
\end{table*}

\section{Empirical Study}
\label{sec:exp}
Here we mainly show that ERAS can improve effectiveness in KG embedding with high efficiency, and provide some insight views.
All codes are implemented with PyTorch \cite{paszke2017automatic} 
and experiments are run on a single TITAN Xp GPU.

\subsection{Experiment Setup}
\label{sec:experimentSetup}

\subsubsection{Data Sets} \label{sec:exp:data}
In our experiments, we mainly conduct experiments on five public benchmarks data sets: 
WN18~\cite{bordes2013translating}, 
WN18RR~\cite{dettmers2018convolutional}, 
FB15k~\cite{bordes2013translating}, 
FB15k237~\cite{toutanova2015observed},
and YAGO3-10~\cite{dettmers2018convolutional},
that have been employed to compare KG embedding models in~\cite{bordes2013translating,kazemi2018simple,liu2017analogical,trouillon2017knowledge,yang2014embedding}. Note that WN18RR and FB15k237 remove duplicate and inverse duplicate relations from WN18 and FB15k, respectively. The statistics of five data sets are summarized in 
Table~\ref{table:KGs}.

\begin{table}[ht]
	\centering
	\caption{Summary of KG benchmark data sets.}
	\label{table:KGs}
	\vspace{-5px}
	\setlength\tabcolsep{4pt}
	\begin{tabular}{c|c|c|c|c|c}
		\hline
		\bf Data set & \bf \#relation & \bf \#entity & \bf \#training & \bf \#validation & \bf \#testing \\ \hline
		WN18     &       18        &    40,943     &     141,442     &       5,000       &     5,000      \\ \hline
		WN18RR    &       11        &    40,943     &     86,835      &       3,034       &     3,134      \\ \hline
		FB15k     &      1,345      &    14,951     &     484,142     &      50,000       &     59,071     \\ \hline
		FB15k237   &       237       &    14,541     &     272,115     &      17,535       &     20,466     \\ \hline
		YAGO3-10   &       37        &    123,188    &    1,079,040    &       5,000       &     5,000      \\ \hline
	\end{tabular}
\end{table}

\subsubsection{Hyperparameter Settings}

The hyperparameters in this work can be mainly categorized into searching and evaluation parameters.
To fairly compare existing scoring functions, including human-designed and searched ones, we search a stand-alone parameter set on the SimplE with the help of HyperOpt, a hyperparameter optimization framework~\cite{bergstra2013making}.
The tuned parameter set includes: learning rate, L2 penalty, decay rate, batch size, embedding dimensions.
Then we compare the stand-alone performance of different scoring functions on the tuned parameter set.
Besides, the searching parameters of ERAS are the number of segments $M$, relation groups $N$, sampled scoring functions $U$ in \eqref{eq:controllerUpdate2} and \eqref{eq:embeddingUpdate}.
Moreover, we optimize embeddings $\bm{\omega}$ with Adagrad~\cite{duchi2011adaptive} algorithm and the controller $\theta$ with Adam~\cite{kingma2014adam} algorithm.

\begin{figure*}[t]
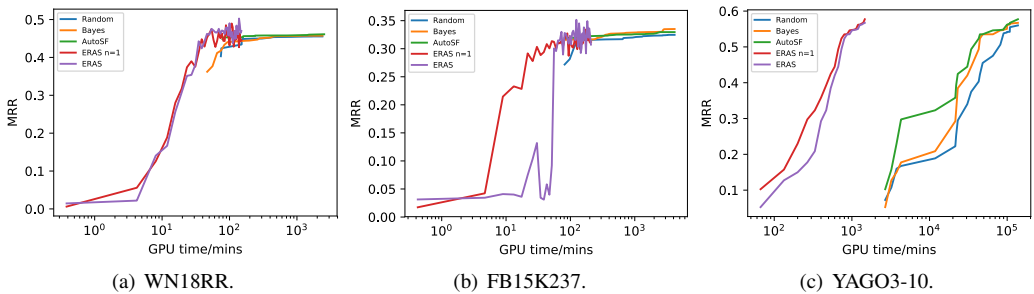

	\centering
	\subfigure[WN18RR.]
	{\includegraphics[width=0.245\linewidth]{Figs/efficiency/wnrr_mrr_time_}}
	\subfigure[FB15K237.]
	{\includegraphics[width=0.25\linewidth]{Figs/efficiency/fb237_mrr_time_}}
	\subfigure[YAGO3-10.]
	{\includegraphics[width=0.245\linewidth]{Figs/efficiency/yago_mrr_time_}}
	
	\vspace{-7px}
	
	\caption{Search efficiency comparison of ERAS with the other popular search algorithms in AutoML.}
	\label{figs:effiency}
\end{figure*}

\subsection{Comparison with KG Embedding Methods}
\label{ssec:expLink}
As in \cite{zhang2019autosf,balazevic2019tucker,zhang2019quaternion}, 
we perform experiments with link prediction and triplet classification task, as they are important testing bed for scoring functions.

\subsubsection{Link Prediction} \label{sec:explp}
We first test the performance of our proposed method on the link prediction task.
This is the test bed to measure KG embedding models and works as an important task in KG completion.
Given the triplet $(h,r,t)\in S_{\text{val}}\cup S_{\text{test}}$, 
the KG embedding model obtains the rank of $h$ through computing the score of $(h',r,t)$ for all entities, and the same for $t$. 
We adopt the classic metrics~\cite{bordes2013translating,wang2014knowledge}: 
\begin{itemize}[leftmargin=*]
\item Mean Reciprocal Ranking (MRR):
$\nicefrac{1}{|S|}\sum_{i=1}^{|S|}\nicefrac{1}{\text{rank}_i}$, where $\text{rank}_i$ is the ranking result; 
and 

\item Hit@1, 
i.e., $\nicefrac{1}{|S|}\sum_{i=1}^{|S|}\mathbb{I}(\text{rank}_i\! \le\! 1)$,
and Hit@10,
i.e., $\nicefrac{1}{|S|}\sum_{i=1}^{|S|}\mathbb{I}(\text{rank}_i\! \le\! 10)$, where $\mathbb{I}(\cdot)$ is the indicator function. 
\end{itemize}

Note that the higher MRR, Hit@1 and Hit@10 values mean better embedding quality. 
We compare the proposed ERAS (Algorithm~\ref{alg:ERAS})
with the popular KG embedding models mentioned in Section~\ref{sec:intro}:
\begin{itemize}[leftmargin=*]
	\item Translational models (TDMs): TransE~\cite{bordes2013translating}, TransH~\cite{wang2014knowledge}, 
	and RotatE~\cite{sun2019rotate};
	
	\item Neural network models (NNMs): NTN~\cite{socher2013reasoning},
	ConvE~\cite{dettmers2018convolutional}, and HypER~\cite{balavzevic2019hypernetwork};
	
	\item 
	Tensor-based models (TBMs): TuckER~\cite{balazevic2019tucker}, 
	HolEX~\cite{xue2018expanding}, 
	QuatE~\cite{zhang2019quaternion}, DistMult~\cite{yang2014embedding}, ComplEx~\cite{trouillon2017knowledge},  
	Analogy~\cite{liu2017analogical} and SimplE~\cite{kazemi2018simple};

	\item The rule-based model: AnyBURL~\cite{meilicke2019anytime};
	
	\item The scoring function search method: AutoSF~\cite{zhang2019autosf}.
\end{itemize}

\begin{table}[t]
\centering
\caption{Hit@1 results for $\text{ERAS}^{N=1}$ and ERAS on the link prediction task at the relation pattern level.}
\label{table:erasRelationAware}
\setlength\tabcolsep{1.5pt}
\vspace{-5px}
\begin{tabular}{l|ccc|ccc}
	\hline
	\multirow{2}{*}{Methods} & \multicolumn{3}{c|}{\bf  Symmetric relations} & \multicolumn{3}{c}{\bf  Anti-symmetric relations} \\ \cline{2-7} 
	&    WN18RR    &   FB15k    &    FB15k237   &      WN18RR    &     FB15k     &     FB15k237     \\ \hline
	Best in Table~\ref{table:relationAware}	&   94.0      & 88.0   &   7.0   &   12.0 & 81.0 & 27.0 \\ \hline
	$\text{ERAS}^{N=1}$	&      93.2   &  86.5  &  5.3       & 11.6   &  80.4 & 26.9   \\ \hline
	ERAS	&     94.3   & 90.0  & 8.8  &  13.2      &    82.1    &    27.9  \\ \hline
\end{tabular}
\end{table}

\begin{table*}[t]
	\caption{Running time analysis of the automated approaches on single GPU in hours.}
	\label{table:runningTime}
	\centering
	\vspace{-5px}
		\begin{tabular}{c|c|c|c|c|c|c||c|c}
			\hline
			\multirow{2}{*}{Methods} & \multicolumn{2}{c|}{\bf AutoSF} &  \multicolumn{2}{c|}{\bf $\text{ERAS}^{N=1}$ }&  \multicolumn{2}{c||}{\bf ERAS }   & \multicolumn{2}{c}{\bf Handed-designed } \\ \cline{2-9}
			& greedy search &   evaluation    &supernet training & evaluation & supernet training &  evaluation   &   DistMult    &          QuatE           \\ \hline
			WN18           & 65.7$\pm$3.0  &   5.5$\pm$0.5 &  3.29$\pm$0.2 & 2.1$\pm$0.1 &   3.54$\pm$0.1    &  2.2$\pm$0.1  &  1.9$\pm$0.1  &       2.0$\pm$0.1        \\ \hline
			FB15K           & 127.1$\pm$5.2 &  20.5$\pm$ 1.3 & 4.55$\pm$0.2 & 19.0$\pm$0.2 &   4.86$\pm$0.2    & 19.49$\pm$0.3 & 8.36$\pm$0.2  &       11.1$\pm$0.4       \\ \hline
			WN18RR          & 38.6$\pm$1.9  &  3.72$\pm$0.6   & 2.97$\pm$0.2 & 0.50$\pm$0.1 &   3.19$\pm$0.1    & 0.52$\pm$0.1  & 0.42$\pm$0.1 &       0.95$\pm$0.1       \\ \hline
			FB15k237         & 61.1$\pm$2.8  &   8.5$\pm$0.4   & 3.22$\pm$0.1 & 4.7$\pm$0.1 &   3.54$\pm$0.1    &  4.8$\pm$0.2  & 2.6$\pm$0.1  &       5.0$\pm$0.3        \\ \hline
			YAGO           & 219.9$\pm$5.1 &   18.9$\pm$2.0    & 17.5$\pm$0.3 & 29.5$\pm$1.1 &   19.8$\pm$0.3    & 30.3$\pm$1.9  & 26.4$\pm$1.5  &       32.6$\pm$2.0       \\ \hline
		\end{tabular}
	\vspace{-10px}
\end{table*}

The comparison of the global effectiveness between ERAS and other methods is in Table~\ref{table:linkPrediction}.
Firstly, it is clear that traditional scoring functions, such as TDMs, NNMs, and TBMs, are not task-aware since no scoring functions can perform consistently over the benchmark data sets. 
This indicates that a single scoring function is hard to adapt to different KGs even though it is a universal scoring function like, as discussed in Table~\ref{table:relationAware}. 
The task-aware method AutoSF can search the KG-dependent scoring functions on five data sets and perform consistently better than traditional scoring functions.
Then, we compare AutoSF with a variant of ERAS, i.e., $\text{ERAS}^{N=1}$, that aims to search a universal scoring function since all relations are assigned into one group (i.e., only task-aware as AutoSF). 
For five benchmark data set, $\text{ERAS}^{N=1}$ shows almost the same performance with AutoSF.
Moreover, as a task-aware and relation-aware method, $\text{ERAS}$ performs better than AutoSF and other manually designed scoring functions.

As discussed in Section~\ref{ssec:motivation}, the existing scoring functions may achieve unsatisfactory performance on specific relation types of certain KG data.	
Corresponding to Table~\ref{table:relationAware}, we investigate the performance of $\text{ERAS}^{N=1}$ and ERAS at the relation type level in Table~\ref{table:erasRelationAware}.
Obviously, the relation-aware method ERAS can consistently achieve outstanding performance on various relation types of any KG data.
Especially, ERAS improves the performance of symmetric relations in FB15k and FB15k237.
However, since the fact of symmetric relation only accounts for 3\% of the test data in FB15k and FB15k237 \cite{rossi2020knowledge}, the global improvement is not so notable.

\subsubsection{Triplet Classification}
To further demonstrate the effectiveness of ERAS, we also conduct the triplet classification experiments on FB15k, WN18RR, and FB15k237, 
where the positive and negative triplets are provided.
We compare our methods with those that have reported results in public papers.
This task aims to answer whether a given $(h,r,t)$ exists or not. In this task, we utilize the accuracy to evaluate the scoring functions. We set the same decision rule of classification in literature~\cite{zhang2019autosf}: predicting a $(h,r,t)$ is positive if $f(h,r,t)>\theta_r$ otherwise negative, where $\theta_r$ is a relation-specific threshold and is inferred by maximizing the accuracy on $S_{\text{val}}$. 
As shown in Table~\ref{table:tripletClassification}, the scoring function searched by relation-aware $\text{ERAS}$ consistently outperforms other BLMs or searched scoring functions.

\begin{table}[ht]
	\centering
	\caption{
		Comparison of the best scoring functions identified by ERAS and the state-of-the-art scoring functions for triplet classification. [$\clubsuit$]: results are taken from \cite{zhang2019autosf}.}
	\label{table:tripletClassification}
	\begin{tabular}{c|c|c|c}
		\hline
		\bf Data set & \bf FB15k & \bf WN18RR & \bf FB15k237  \\ \hline
		DistMult$^\clubsuit$    &    80.8   &  84.6   & 79.8   \\ \hline
		Analogy$^\clubsuit$   &    82.1   &  86.1   & 79.7 \\ \hline
		ComplEx$^\clubsuit$    &  81.8   & 86.6  & 79.6 \\ \hline
		SimplE$^\clubsuit$  &  81.5     &   85.7 & 79.6 \\ \hline
		AutoSF$^\clubsuit$  &  82.7      & 87.7 & 81.2 \\ \hline
		$\text{ERAS}$ &   \bf 82.9     &  \bf 88.0&  \bf 81.4\\ \hline
	\end{tabular}
\end{table}

\begin{figure*}[ht]
	\centering
	\subfigure{\includegraphics[width=1.0\linewidth]{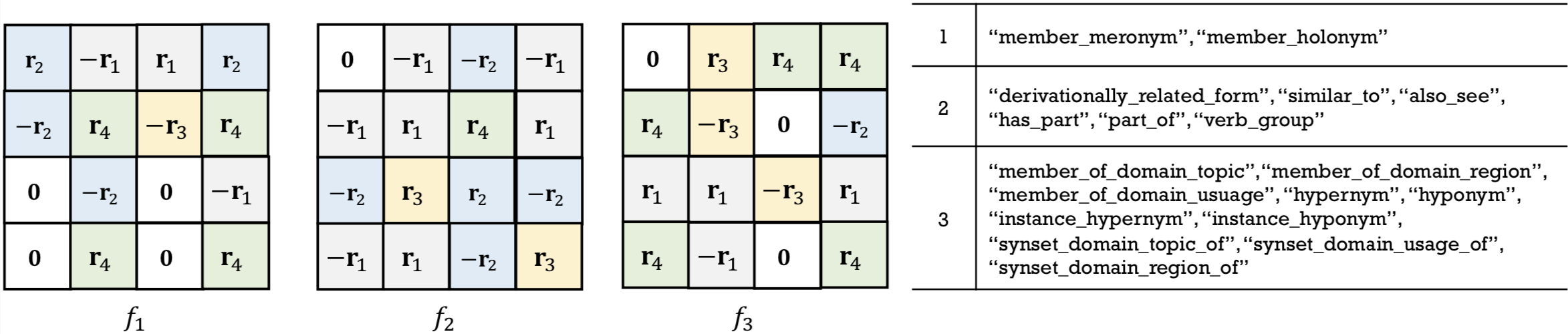}}
	\vspace{-10px}
	\caption{The example of searched relation-aware scoring functions by ERAS on WN18.}
	\label{figs:caseSFswn18}
\end{figure*}
\vspace{-5px}

\begin{figure*}[ht]
	\centering
	\subfigure{\includegraphics[width=1.0\linewidth]{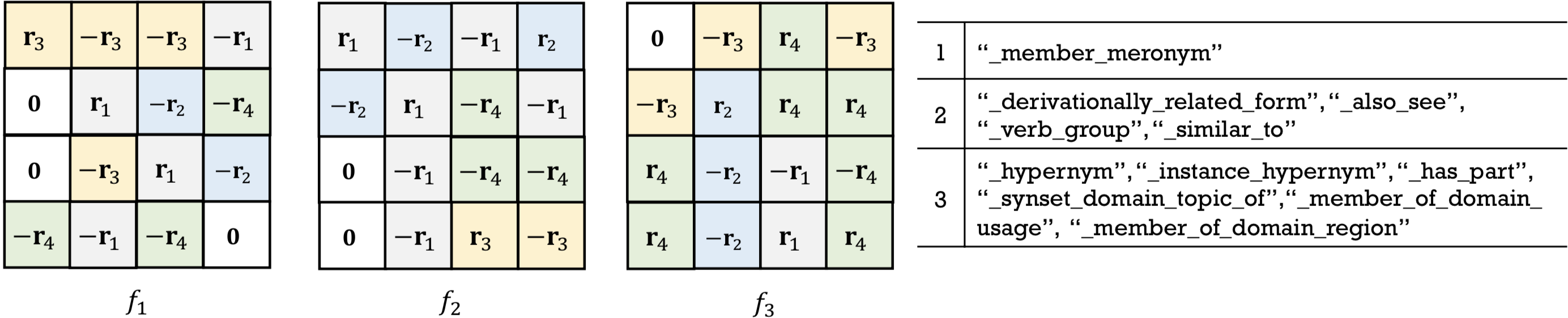}}
	\vspace{-10px}
	\caption{The example of searched relation-aware scoring functions by ERAS on WN18RR.}
	\label{figs:caseSFswn18rr}
\end{figure*}

\subsection{Comparison with AutoML Search Methods}
\label{ssec:com:automl}
ERAS enables embedding sharing to improve the scoring function search efficiency.
Hence we also compare the scoring function search efficiency of ERAS and $\text{ERAS}^{N=1}$ with other automated search algorithms, i.e. 
AutoSF \cite{zhang2019autosf}, random search~\cite{li2019random}, and Bayes algorithm~\cite{bergstra2013making}, over three benchmark datasets. 
As shown in the Figure~\ref{figs:effiency}, both ERAS and $\text{ERAS}^{N=1}$ complete the search very quickly.
That is because other AutoML methods have to train hundreds of candidate scoring functions to convergence, while ERAS and $\text{ERAS}^{N=1}$ avoids the time-consuming training by one-shot way. 
Compared with $\text{ERAS}$, $\text{ERAS}^{N=1}$ convergess faster in search procedure since it does not dynamically assign relations and search corresponding scoring functions for different groups.
It is worth noting that ERAS has unstable performance at beginning of the search. 
That is because FB15k237 has much more relations than another 2 data sets.
It takes some time to find proper relation assignments at the start.
Furthermore, other automated methods can achieve higher performance than ERAS in the search strategy.
ERAS consistently updates the candidate scoring functions in every mini-batch data, 
which results in that the searched scoring functions cannot be well trained with one or several mini-batches.  
But other automated methods train the searched scoring functions with all training data until convergence. 

We also summarize the running time on the five data sets in Table~\ref{table:runningTime}. 
AutoSF sets the embedding dimension $d$ to 64 for all data sets, while ERAS enables much faster search and hence sets $d=512$.
The larger dimensionality enables us to evaluate the scoring function more accurately.
As shown in Table~\ref{table:runningTime}, the search time of AutoSF is significantly reduced by $\text{ERAS}^{N=1}$.
But recall that the effectiveness of $\text{ERAS}^{N=1}$ is the same with AutoSF in Table~\ref{table:linkPrediction}.
This indicates that $\text{ERAS}^{N=1}$ can extremely shorten the task-aware search time but maintain the same effectiveness with AutoSF.
Moreover, although ERAS searches from a larger search space that is relation-aware, it reduces the search time of AutoSF by one order with a 	large dimension size $d=512$ and improves effectiveness as in Table~\ref{table:linkPrediction}.
In summary, ERAS can more efficiently search for more effective scoring functions.


\subsection{Case Study: The Searched scoring functions}
To show the searched scoring functions by ERAS are relation-aware, 
we use searched scoring functions from WN18 and WN18RR
as examples and plot them in Figure~\ref{figs:caseSFswn18} and ~\ref{figs:caseSFswn18rr}.
As we can see, the three searched scoring functions have distinct patterns and are relation-aware.
Moreover, the relations are grouped into general asymmetry, symmetry, and anti-symmetry. 
And the three SFs searched have distinct patterns and can handle their corresponding relations.

\subsection{Ablation Study}
To investigate the influence of different components of ERAS, we conduct several ablation studies.
	
\subsubsection{Impact of Evaluation Measurement and Optimization Algorithm}
\label{sssec:ablationEvaluate}
\begin{figure}[t]
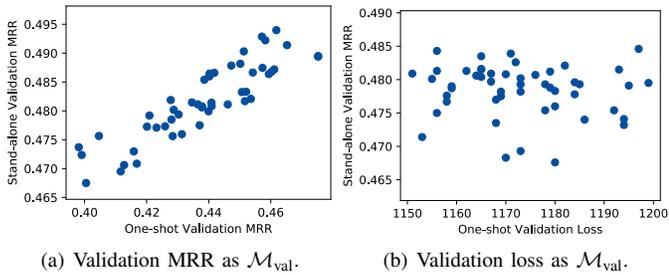

	\centering
	\subfigure[Validation MRR as $\mathcal{M}_{\text{val}}$.]{\includegraphics[width=0.49\linewidth]{Figs/ablation/correlation_}}
	\subfigure[Validation loss as $\mathcal{M}_{\text{val}}$.]
	{\includegraphics[width=0.49\linewidth]{Figs/ablation/correlation_loss_}}
	\caption{The correlation between stand-alone validation MRR with different $\mathcal{M}_{\text{val}}$ settings of various searched scoring functions on WN18RR.}
	\label{figs:ablationCorrelation}
\end{figure}

As discussed in Section~\ref{ssec:comp:oas}, the deep and complex supernet design will probably lead to the biased evaluation problem. 
In Figure~\ref{figs:ablationCorrelation}(a), we first demonstrate the correlation between stand-alone validation MRR with one-shot validation MRR (i.e., $\mathcal{M}_{\text{val}}$) of various scoring functions in ERAS. 
It is obvious that one-shot validation MRR has near positive correlation with the stand-alone validation MRR. 
Therefore, the simplified design of supernet makes embedding sharing work, and there is no biased evaluation problem.

To further investigate the impact of $\mathcal{M}_{\text{val}}$ and 
optimization algorithms, we compare ERAS with following variants:
\begin{itemize}[leftmargin=*]
	\item $\text{ERAS}^{\text{los}}$ utilizes the validation loss $\mathcal{L}$ to replace $\mathcal{M}_{\text{val}}$.
	Other steps are same with ERAS.
	
	\item 
	$\text{ERAS}^{\text{dif}}$ first replaces $\mathcal{M}_{\text{val}}$ with $\mathcal{L}$ as $\text{ERAS}^{\text{los}}$ does.
	Then the differentiable measurement $\mathcal{L}$ enables the differentiable optimization algorithm~\cite{liu2018darts,yao2019differentiable} for search.
	The detailed implementing $\text{ERAS}^{\text{dif}}$ is presented in Appendix.
\end{itemize}

We show the correlation between stand-alone validation MRR with different $\mathcal{M}_{\text{val}}$ settings in Figure~\ref{figs:ablationCorrelation}.
Obviously, compared with using MRR as $\mathcal{M}_{\text{val}}$ in Figure~\ref{figs:ablationCorrelation} (a), Figure~\ref{figs:ablationCorrelation} (b) shows that using $\mathcal{L}$ as $\mathcal{M}_{\text{val}}$ has a low correlation with stand-alone validation MRR.
This indicates that validation loss in search strategy cannot well evaluate the stand-alone performance of scoring functions.
Subsequently, in Table~\ref{tab:ablation}, we can observe that the performance of ERAS using MRR as $\mathcal{M}_{\text{val}}$ is better than that of two variants, i.e., $\text{ERAS}^{\text{los}}$ and $\text{ERAS}^{\text{dif}}$.
Moreover, $\text{ERAS}^{\text{los}}$ is worse than $\text{ERAS}^{\text{dif}}$ because optimizing loss with the RL approach could not make use of the differentiable nature of $\mathcal{L}$.

\begin{figure}[t]
\centering
\subfigure[WN18RR.]{\includegraphics[width=0.49\linewidth]{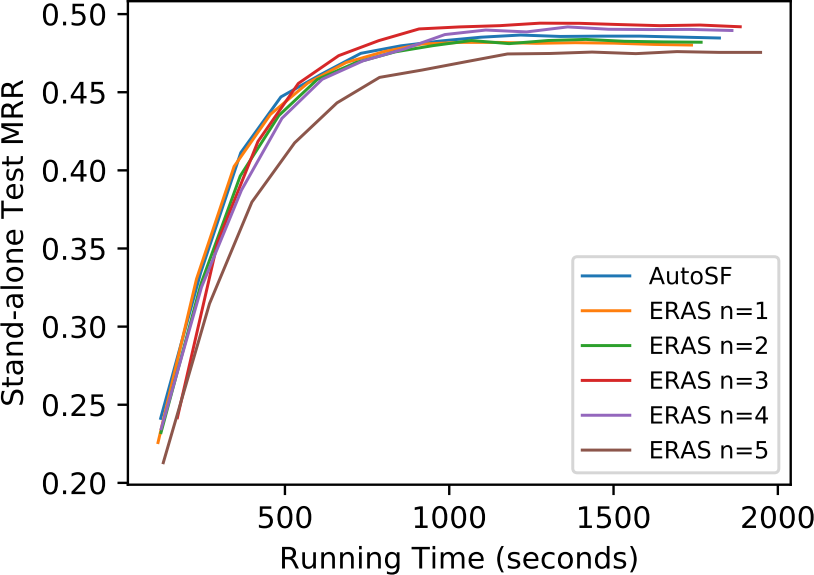}}
\subfigure[FB15k237.]{\includegraphics[width=0.48\linewidth]{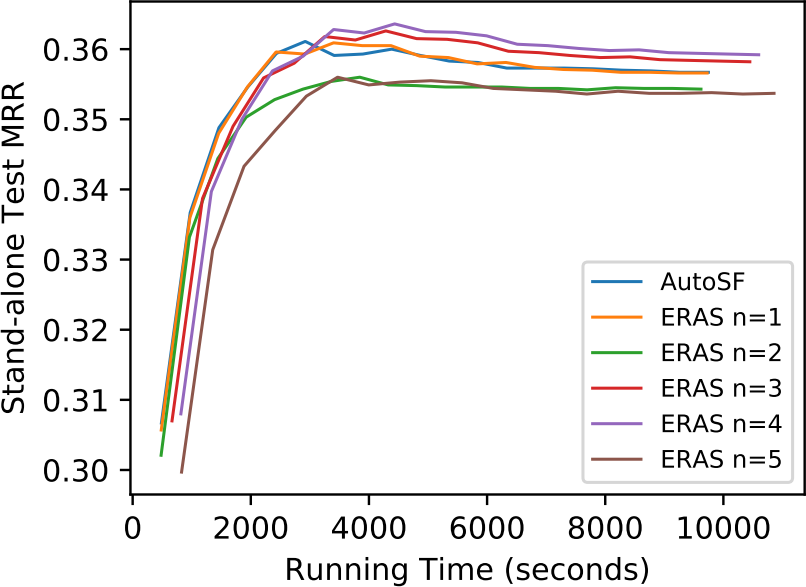}}
\caption{Comparison on time (sec) of model training vs. testing MRR with different number of groups $N$ in ERAS.}
\label{figs:ablationGroup}
\end{figure}

\begin{table}[ht]
\caption{Comparison of the variants of ERAS on the link prediction task.}
\label{tab:ablation}
\setlength\tabcolsep{3pt}
\centering
\begin{tabular}{C{25px}|c|c|c|c|c|c}
	\hline
	\multirow{2}{*}{Section} &\multirow{2}{*}{Variant} &      \bf WN18    &    \bf WN18RR   &    \bf FB15K    &  \bf FB15k237  &     \bf YAGO3-10  \\ \cline{3-7}
	&&        MRR       &        MRR          &        MRR       &        MRR           &        MRR        \\ \hline
	
	\multirow{2}{*}{\ref{sssec:ablationEvaluate}}&$\text{ERAS}^{\text{los}}$         & 0.944   &  0.485 &  0.840   &  0.344& 0.560  \\
	
	&$\text{ERAS}^{\text{dif}}$         &  0.949     &  0.485 &   0.848   &  0.355 &   0.565   \\\cline{1-7}
	
	\ref{sssec:ablationOptim}&$\text{ERAS}^{\text{sig}}$         &  0.945      &  0.480  &   0.844   &  0.338 &   0.559  \\\cline{1-7}
	
	\multirow{2}{*}{\ref{sssec:ablationGroup}}&$\text{ERAS}^{\text{pde}}$         &   0.950    &  0.489   &   0.850   &   0.349   & 0.570   \\
	
	&$\text{ERAS}^{\text{smt}}$   &  0.948   &  0.485  & 0.845  &  0.347 & 0.565   \\\cline{1-7}
	
	&$\text{ERAS}$  &     \bf 0.953      &     \bf 0.492      &    \bf{0.855}       &     \bf 0.365      &     \bf 0.577       \\ \hline
\end{tabular}
\end{table}

\subsubsection{Impact of Optimization Level} 
\label{sssec:ablationOptim}
In this paper, Definition~\ref{def:problem} formulates the problem with a bi-level optimization objective.
As stated in Section~\ref{ssec:prodef}, bi-level optimization can benefit ERAS by updating the scoring functions and embeddings separately.
To investigate the impact of optimization level, we add another variant of ERAS as $\text{ERAS}^{\text{sig}}$, which utilizes the training set to update \eqref{eq:objective} in Definition~\ref{def:problem} (i.e., single-level problem).
In Table~\ref{tab:ablation}, compared with $\text{ERAS}^{\text{sig}}$, $\text{ERAS}$ demonstrates the bi-level optimization is needed because optimizing scoring functions on the validation set encourages ERAS to select scoring functions that generalize well rather than scoring functions that overfit on the training data.

\subsubsection{Impact of Grouping Approaches} 
\label{sssec:ablationGroup}
To explore more about the influence of different grouping approaches,
we set two variants of ERAS as follows:
\begin{itemize}[leftmargin=*]
	\item $\text{ERAS}^{\text{pde}}$ does not update $\bm{B}$ during search.
	Instead, it fixes groupings based on the embedding trained from SimplE.
	\item $\text{ERAS}^{\text{smt}}$ groups relations based on their semantic meanings (i.e., symmetric, anti-symmetric, asymmetric and inverse).
\end{itemize}

We compare these two variants with ERAS as shown in Table~\ref{tab:ablation}.
Generally, the performance of $\text{ERAS}^{\text{smt}}$
is unsatisfactory due to the bias between human understanding of relation groups with the proper groups derived from data.
In short, the performance comparison also indicates the importance of dynamically assigning relation groups in the search strategy, which encourages relations to be assigned to the appropriate scoring function.

\subsubsection{Impact of Grouping Numbers} 
\label{sssec:exp:group}

To investigate the impact of grouping numbers in ERAS,
we summarize the performance of searched scoring functions by AutoSF and different settings of ERAS in
Figure~\ref{figs:ablationGroup}, i.e., $\text{ERAS}^{N}$ ($N\in\{1,2,\cdots,5\}$) on WN18RR and FB15k237.
Generally, the more groups there are, the longer the running time is.
And ERAS achieves the best performance when $N=3$ or $4$.
Comparing AutoSF and $\text{ERAS}^{N=1}$, they have a similar learning curve since their model complexities are the same.

\subsubsection{Impact of Block Numbers}
\begin{figure}[t]
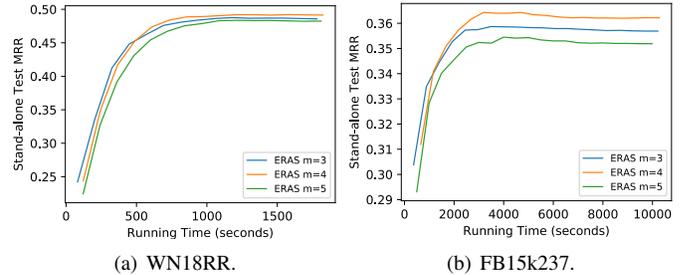

	\centering
	\subfigure[WN18RR.]{\includegraphics[width=0.49\columnwidth]{Figs/ablation/ablation_seg_wn18rr_}}
	\subfigure[FB15k237.]{\includegraphics[width=0.49\columnwidth]{Figs/ablation/ablation_seg_fb15k237_}}
	\caption{Comparison on time (sec) of model training vs. testing MRR with different number of blocks $M$ in ERAS.}
	\label{figs:caseSeg}
\end{figure}

AutoSF fixes the block number $M$ to 4 due to the efficiency issue.
Once $M$ is changed (e.g., $M=3$ or $M=5$), all design details in AutoSF must be re-done.
On the contrary, the efficiency of ERAS allows more flexible settings of $M$.
Here we try $M\in \{3,4,5\}$ in order to learn more about how the block number $M$ influences the ERAS performance. As shown in Figure~\ref{figs:caseSeg}, we can observe that $M=4$ does have excellent performance among $\{3,4,5\}$.

\section{Conclusion}
\label{sec:conclude}

In this paper, we propose a new automated machine learning (AutoML) method for designing scoring functions (scoring functions) in knowledge graph embedding.
First, we design a relation-aware search space, which is motivated by our analysis of how existing scoring functions adapt to different relations.
Then, we represent the new search space as a supernet in the form of a graph and propose to search through the supernet by one-shot architecture search methods. 
Experimental results on benchmark data sets well demonstrate not only the efficiency of our approach but also the competitive effectiveness.

For future works,
one interesting direction to connect ERAS with graph neural network~\cite{zhou2018graph};
another direction worth to try is utilizing path instead of triplet to exploit higher-order information in KGs~\cite{zhang2019neural}.

\section{Acknowledgements}
This work is partially supported by National Key Research and Development  Program of China Grant no. 2018AAA0101100, the Hong Kong RGC GRF Project 16202218 , CRF Project C6030-18G, C1031-18G, C5026-18G, AOE Project AoE/E-603/18, China NSFC No. 61729201, Guangdong Basic and Applied Basic Research Foundation 2019B151530001, Hong Kong ITC ITF grants ITS/044/18FX and ITS/470/18FX, Microsoft Research Asia Collaborative Research Grant, Didi-HKUST joint research lab project, and Wechat and Webank Research Grants.


\bibliographystyle{ieeetr}
\bibliography{sample-base}

\appendix


\section*{Details of Implementing $\text{ERAS}^{\text{dif}}$}
\label{appendix:erasDif}
We propose a supernet view of the scoring function search space as in \eqref{eq:sup}.
This supernet design allows us to employ differentiable OAS methods when we use the loss $\mathcal{L}$ as $\mathcal{M}_{\text{val}}$.
Following NASP~\cite{yao2019differentiable}, $\text{ERAS}^{\text{dif}}$ can update the architecture weight $\bm{A}$ by gradient descent 
as:
\begin{align*}
\bm{A} \leftarrow \bm{A} - \epsilon
\sum\nolimits_n
\sum\nolimits_{(h,r,t)\in S_{\text{val}}}
\nabla_{\bm{A}}
B_{rn}\cdot
\ell(f_n(\bm{h},\bm{r},\bm{t})).
\end{align*}
Then \eqref{eq:objective} can be optimized by above equation. Other steps of $\text{ERAS}^{\text{dif}}$ are same with ERAS.


\end{document}